\documentclass[letterpaper, 10 pt, conference]{ieeeconf}
\usepackage{algorithm}
\usepackage[noend]{algpseudocode}
\usepackage{adjustbox}
\usepackage{booktabs}
\usepackage{acronym}
\usepackage{amsmath, amssymb}
\newcommand{\etal}{et al.~}
\newcommand{\ie}{\textit{i}.\textit{e}.~}

\newcommand{\bDAD}{binary DAD-Net}
\newcommand{\BDAD}{Binary DAD-Net}
\usepackage{amssymb}
\usepackage[thinc]{esdiff}
\usepackage{pifont}
\newcommand{\cmark}{\ding{51}}
\newcommand{\xmark}{\ding{55}}
\usepackage{url}
\usepackage{array}
\newcolumntype{P}[1]{>{\centering\arraybackslash}p{#1}}

\usepackage{xcolor}
\usepackage{multirow}
\usepackage{subfigure}
\usepackage[capitalise]{cleveref}

\newcommand{\sign}{\texttt{sign}}
\DeclareMathOperator{\E}{\mathbb{E}}
\IEEEoverridecommandlockouts                              % This command is only needed if 
\title{\LARGE \bf Binary DAD-Net: Binarized Driveable Area Detection Network for Autonomous Driving
}

\author{Alexander Frickenstein$^{*1}$, 
Manoj-Rohit Vemparala$^{*1}$,
Jakob Mayr$^{*1}$,\\
Naveen-Shankar Nagaraja$^{1}$, 
Christian Unger$^{1}$,  
Federico Tombari$^{2,3}$, 
Walter Stechele$^{2}$
\thanks{*Authors contributed equally}% <-this % stops a space
\thanks{$^{1}$BMW Group, Autonomous Driving, Munich, Germany,} \thanks{{\tt\small<Firstname>.<Lastname>@bmw.de}}
\thanks{$^{2}$Technical University of Munich, Munich, Germany,}
\thanks{{\tt\small tombari@in.tum.de},\hspace{0.5mm} {\tt\small walter.stechele@tum.de}} 
\thanks{$^{3}$Google Inc. , Zurich, Switzerland}%
}

\begin{document}

\maketitle
\thispagestyle{empty}
\pagestyle{empty}

\begin{abstract}
Driveable area detection is a key component for various applications in the field of autonomous driving (AD), such as ground-plane detection, obstacle detection and maneuver planning. 
Additionally, bulky and over-parameterized networks can be easily forgone and replaced with smaller networks for faster inference on embedded systems.
The driveable area detection, posed as a two class segmentation task, can be efficiently modeled with slim binary networks. This paper proposes a novel \textit{binarized driveable area detection network (\bDAD)}, which uses only binary weights and activations in the encoder, the bottleneck, and the decoder part. The latent space of the bottleneck is efficiently increased (\textbf{$\times$}32\textbf{$\rightarrow$}\textbf{$\times$}16 downsampling) through binary dilated convolutions, learning more complex features.
Along with automatically generated training data, the \bDAD{} outperforms state-of-the-art semantic segmentation networks on public datasets. In comparison to a full-precision model, our approach has a \textbf{$\times$}14.3 reduced compute complexity on an FPGA and it requires only 0.9MB memory resources. 
Therefore, commodity SIMD-based AD-hardware is capable of accelerating the \bDAD.
%Convolutional Neural Networks (CNNs) have demonstrated high-quality results in Semantic Segmentation tasks. However, tasks such as Binary ground plane detection suffer from over parametrization and large amounts of redundancy, which results in inefficient inference compuation. This paper introduces a novel binary ground plane detector, just using binary weights and activations. The binary ground plane detector is highly suitable for mobile applications like those in the field of autonomous driving, with restricted computational resources and limited power capabilities. The proposed binary model incorporates the used of automatically generated training data with numerous contracting pixel labels. But these data allow a low price and swift adaptation to the environment and the sensor setup. Thus the binary ground plane detector, trained on automatically generated training data, demonstrate the superior performance over conventional semantic segmentation models in the field, when it comes to a deployment on an embedded hardware. DAD-Net is \(xx.x \times\) less memory resources and \(xx.x \times\) less MAC-operations.
\end{abstract}

\section{Introduction}

%The drivable area of an autonomously operating ground agent is defined as the ground plane which can be driven on and is free of obstacles.
% \ale{it took 1 week to fix the previous intro}
Artificial Intelligence is often seen as the key enabler for fully autonomous driving due to the recent unprecedented success of deep learning (DL). However, two other key factors also need to be addressed in the field of AD. 
%A resource constrained deployment needs to address three more factors: (1) a safe modularized software (2) an efficient real-time application. 
% \nsn{in the current storyline, the motivation to pick DAD is not inferred. But rather it gets already stated that we are using DAD. The changes I had done exactly had that, which are now removed (instead of commenting out). Since, this is not git it would be nice if changes are commented out (removing one or two lines is ok).}
%Concurrent to this process, a novel dedicated two class binary drivable area detector network (\bDAD) is introduced significantly reducing the memory demand and the computational complexity.

\begin{figure}
    \centering
    \subfigure[Manually labeled data.]{\includegraphics[width=4cm]{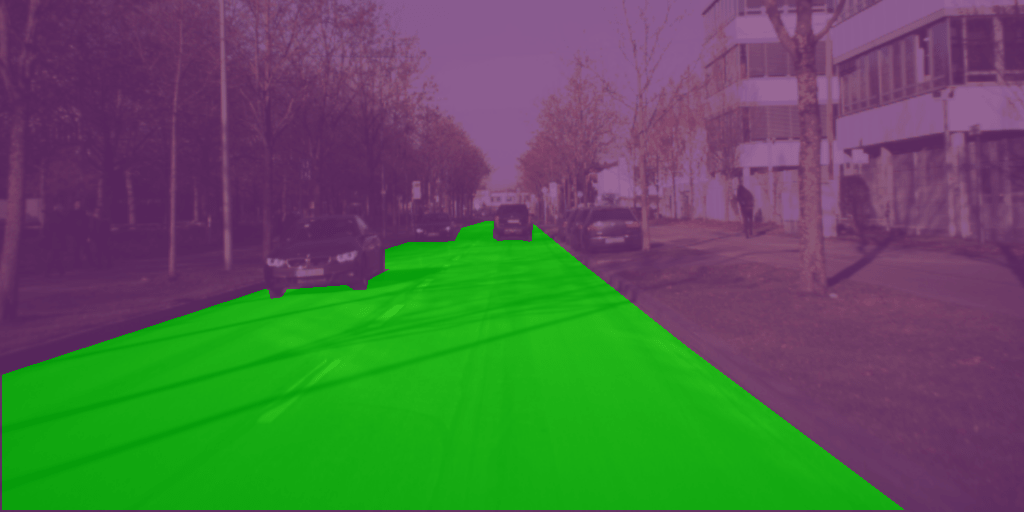}
    \label{fig:third_sub}}
    \subfigure[Prediction of \bDAD.]{\includegraphics[width=4cm]{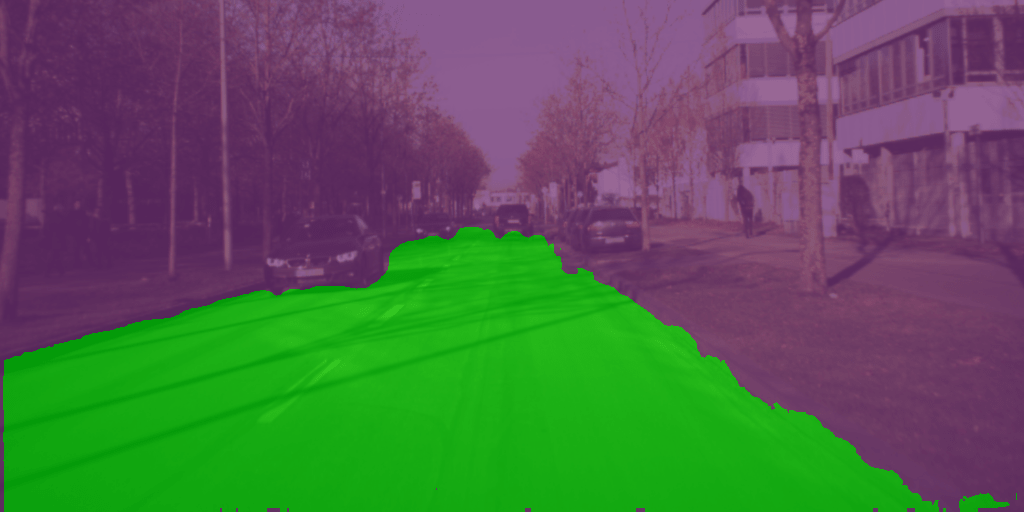}
    \label{fig:third_sub}}
    \caption{Segmentation output of \bDAD{} using only $\pm 1$ weights and activations on proprietary fleet data. The misclassified driveable area pixels in (b) can be easily regularized by a groundplane detection algorithm.}
\label{fig:teaser}
\end{figure}

\noindent\textbf{Verifiable Software: }With safety in mind, the AD stack is often modularized ~\cite{modularAD} into sensing and mapping, perception, and (path) planning blocks. 
Adhering to the modularization nature of AD, we pose the driveable area detection (DAD) as a two class segmentation task. 
In view of the DAD task, the segmentation task for topologically open contours, like roads, is relatively easier than object segmentation where missing a part can have adverse effects, see Fig.~\ref{fig:teaser}.
Moreover, compared to a multi-class segmentation the two class driveable area detection offers a higher precision due to better separability between their respective features.
% \nsn{there is absolutely no relation between automatically generated labels and resource constrained deployment, doesn't fit to the storyline. Hence, commented out. TDG should only be in the experimental part}
%\noindent\textbf{End-to-End Training: }Second, training semantic segmentation networks in an end-to-end manner requires a large amount of costly training data. Hand-labeled ground-truth annotations are mostly expensive and limited to specific driving scenarios.
%Automatic annotations allow to quickly migrate the training data to technical and environmental domain shifts avoiding costly hand-labeled data \cite{Mayr.01.10.201805.10.2018}.

\noindent\textbf{Efficient Application: }Secondly, AD relies on a real-time system which implicitly imposes resource constraints (memory, bandwidth, run-time) on the underlying algorithms.
Due to the real-time requirement, the performance of CNNs also need to be measured w.r.t. the power consumption (apart from standard metrics).
% \nsn{pls dont add random `the's. }\ale{it is not readable without! Sentences make no sence ... dont remove necessary ones!} 
Recent literature studies lightweight CNNs based on network optimization, i.e. pruning~\cite{DSC} and quantization~\cite{WinoCNN}. 
One extreme is represented by Binary Neural Networks (BinaryNets) by constraining weights and activations to $\pm 1$. 
%BinaryNets are challenging to train as compared to their fixed and floating point counterparts mainly due to a huge reduction in trainable parameters and non-differentiable gradients arising due to binarization \cite{bnn_plus}. 
Computationally, they use SIMD-based logical gate (bit-wise) operations instead of  full-precision multiply-accumulate (MAC) operations. BinaryNets, are light weight w.r.t. the memory demand and computational cost, offer a trade-off between efficiency vs. accuracy. 

Considering these two enabling factors for autonomous driving systems naturally make the two class driveable area detection~\cite{seg_pyramid_jakob} a perfect case for BinaryNets significantly reducing the memory demand and the computational complexity. To best of our knowledge \bDAD{} is the first work fully binarizing a semantic segmentation model for the DAD task.
The contribution of this work are summarized as follows:
% \man{todo: rework on this contribution points. Some how gives an impression of just merging domains in the first point}
\begin{itemize}
% \item Introducing novel binarized Bottleneck to the Drivable Area Detector and show superiority to SotA binary Bottleneck (BPAC).
\item Efficiently optimized various blocks in the model's encoder, bottleneck and decoder, combining structural and local binary approximation schemes. A detailed ablation study is provided investigating different variants in \bDAD{}.
\item The proposed binary model performs similar to the full-precision network gaining $14.3 \times$  computational efficiency and $15.8 \times$ memory saving for Cityscapes dataset on the DAD task. 
% \nsn{this isnt a contribution, it is just an application. The real contribution is the second one,and the configuration exploration}\ale{;)}
% \item Binarizing all the convolutions of binary DAD-Net (also binarize the decoder), by not using transpose convolutions.
\item The performance of \bDAD{}  is increased when pre-trained on automatic annotations.
\end{itemize}

\section{Related work}
%\nsn{introduction about each section is not required in research papers. thats more of a talk/powerpoint style (showing outlines). -- its a waste of space adding not much value. if the subsection names have clarity then the outline is automatically clear..this is what you can often see in high quality Tier-1 papers. For Journal papers its perhaps ok to add a intro to each section, bcos journal papers are long and have no page limit.}\ale{8 Reviews pointed to to introduce a section to make it readable esecially as all the thes and as were missing!} \nsn{I wouldn't mind having a word with those 8! besides, I am pretty much aware of where a `the' should be added. I would suggest creating a mattermost channel and chatting over there. Lets not clutter the paper} \ale{WRONG: NIPS Paper/Tire1 introduces section: 3  Binarization methods In this section, we detail our binarization method, which is termed ABC-Net (Accurate-Binary-Convolutional) for convenience. Bear in mind that during }
In the literature, several methods have been proposed to address the task of semantic segmentation, see Sec.~\ref{sec:related_semseg}. Efficient training and perception, \ie binary neural networks, are detailed in Sec.~\ref{sec:related_bnn}.

\subsection{Driveable Area Detection}
\label{sec:related_semseg}

%An accurate localization of the drivable area is crucial for a safe and comfortable path and trajectory planning. 
The task of detecting the driveable area from images has been extensively studied. 
Monocular camera systems \cite{ Lourakis.1997} use a homography computed from two consecutive images which provides information about the ground plane. Similarly, \cite{Yao.2015} utilize a single camera and a variety of features including the homography in combination with a SVM learning approach to solve the task. Early systems focused only on the driveable area, whereas current deep learning approaches usually address the problem of full-image segmentation which includes the driveable area. 
%Classical model driven approaches showed that it is highly beneficial to utilize the depth information, e.g. from stereo cameras, to detect the ground plane. The \emph{v-disparity map}~\cite{Labayrade2002} introduced a linear model fitting of the drivable area under the assumption of a ground vehicle. This principle was adopted by numerous subsequent works \cite{Soquet.2007,Zhao.2007,Yiruo.2013,Dekkiche.2016}.
%Semantic segmentation has made progress in the recent years with DL. 
One of the first prominent semantic segmentation models proposed was the Fully Convolutional Network (FCN), is successfully adopted by Shelhamer \etal~\cite{Shelhamer:2017:FCN:3069214.3069246}.
It is shown that these networks are difficult to train from scratch and require a pre-trained classification model (encoder).
%This is refereed as encoder which uses a combination of convolution and pooling operation to extract low level features.
Another important aspect of FCN are the skip connections which capture the intermediate features from the high level feature maps during the up-sampling stage.
This method paved path to further, more structured models such as UNet~\cite{RFB15a}. %The decoding stage of UNet up-samples feature maps using transposed convolution for each subsequent down-sampling stage. The up-sampled feature maps are fused with the corresponding features maps from the encoder (e.g VGG16) with the same resolution. 
This structured up-sampling provides higher accuracy than single $\times8$ up-sampling, i.e. FCN. However, this increases the computational complexity.
%The fusion method proposed in the framework is concatenation. 
DeepLab, proposed in~\cite{deeplabv3plus2018}, utilizes dilated convolution instead of down-sampling the feature maps maintaining the sufficient receptive field. The pooling or strided convolution is avoided for the last set of feature maps. This would increase the computational costs as the convolution is performed on larger feature maps. The encoder network is downsampled by a factor of 8/16 instead of 32. The down-sampled featured maps are then passed to a spatial pyramid pooling module, which consists of parallel dilated convolution with different rates followed by concatenation and point-wise convolution. This module produces better segmentation results by extracting multi-scale information. 
Multi-class semantic segmentation has a negative effect on the precision of the driveable area detection algorithm and their vast number of MAC operations making the application impractical for embedded systems.

\subsection{Binary Neural Networks}
\label{sec:related_bnn}
Binarization of CNNs attempts to constrain weights and activations to just $\pm 1$. However, Binary Neural Networks observe a degradation in accuracy compared to their full-precision counterpart. 
BinaryNets proposed by Hubara \etal\cite{NIPS2016_6573} relies on deterministic binarization functions and the STE estimator \cite{Bengio2013EstimatingOP} during training.
%, \eg $x^b=\texttt{sign}(x)$. 
%The gradient of the sign operation diminishes everywhere and thus the full precision weights during training are updated using STE estimator proposed by Bengio \etal\cite{Bengio2013EstimatingOP}. 
The degradation in accuracy of full-precision to binary weights can be reduced by suitable approximation techniques. In the binary model XNOR-Net,
introduced by Rastegrati \etal\cite{rastegariECCV16}, the real-valued weights and activations are estimated by introducing scaling factors alongside with the binary weights and activations. 
CompactBNN, proposed by Tang \etal\cite{Tang2017HowTT}, focuses on improving the approximation towards the activations, as they observe that binarizing activations is more challenging compared to the weights.
\cite{Tang2017HowTT} also propose the trainable parametric ReLU as activation function to further improve the training.
Recently, ABC-Net~\cite{NIPS2017_6638} projects both, the full-precision weights and activations into corresponding linear combinations of its binary approximation with individual shifting and scaling factors.
Further, they argue that a BinaryNets with multiple binary weight and activation bases is more suited for embedded systems than an equivalent fixed-point quantized CNN. The MAC operation consumes $>8\times$ more power than a bit-wise operation using 45nm CMOS technology \cite{Han:2015:LBW:2969239.2969366}. 
All publications mentioned above have consequently improved BinaryNets for image classification. Binary object detection models are studied by Hanyu \etal\cite{Sun2018FastOD}.
Zhuang \etal\cite{CVPR19Zhuang3} extend the approximations further towards the structure level and proposes GroupNet with multiple binary bases. 
To the best of our knowledge this is the only work in the domain of BinaryNets reporting results to semantic segmentation. 
GroupNet introduces the Binary Parallel Atrous Convolution (BPAC) module. 
The BPAC module consists of multiple dilated convolutions with various dilation rate (up to 16), which causes irregular memory accesses (inefficient) and an higher power-consumption of the memory controller \cite{DSC}.
Moreover, introducing multiple binarizations indices in the \bDAD{} is not beneficial for the DAD task, as discussed in Tab.~\ref{tab:ablation_binary_dad_binarization} of the experimental section.

\section{Binary Drivable Area Detection Network}
\label{sec:binary_dad}
\begin{figure*}[ht]
\centering
  \includegraphics[width=0.95\textwidth, angle=0]{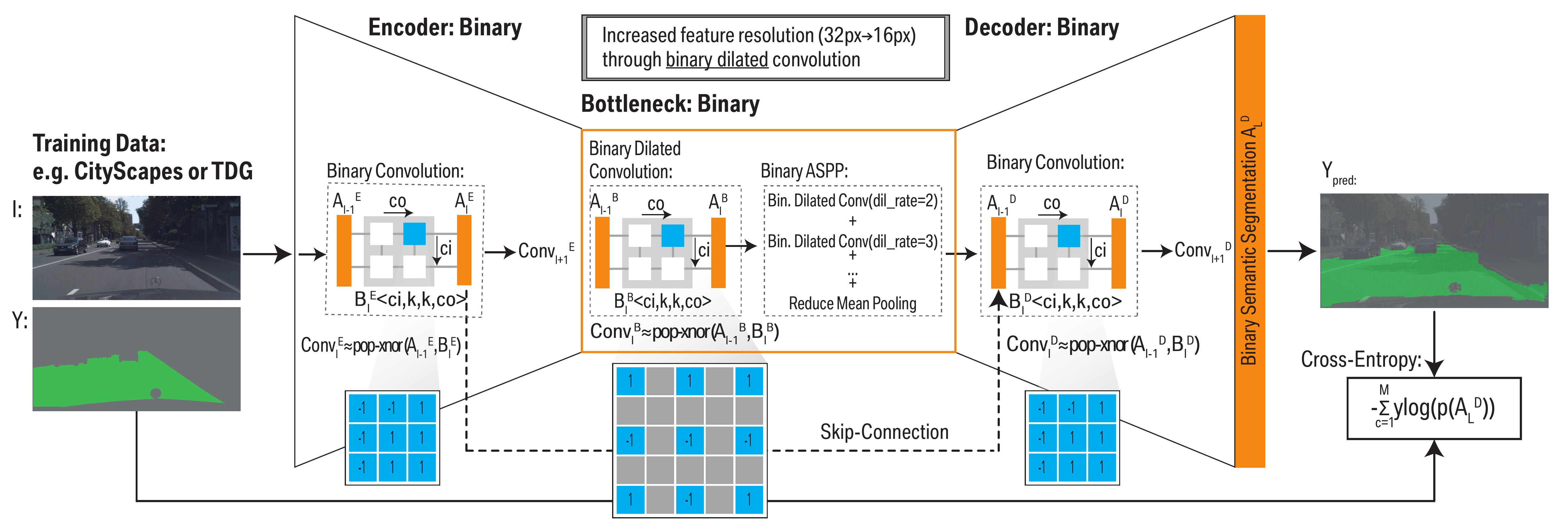}
	\caption{Overview of binary DAD-Net. The binarized network consists of three parts, namely an encoder, a bottleneck and a decoder. Binary dilated convolution in the bottleneck ensures an extended feature resolution. The feature response is efficiently increased from 32 to 16 neurons through binary dilated convolutions.}
	\label{fig:binarynet}
\end{figure*}
The proposed driveable area detector is inspired by autoencoder-based networks with skip connections, \ie DeepLabV3~\cite{deeplabv3plus2018}. As the name implies, \bDAD{} has binary representations in all three parts of the model: the encoder, bottleneck (latent space) and decoder. The modules are detailed in Sec.~\ref{subsec:BinaryEncoder}-\ref{subsec:BinaryDecoder}. \BDAD{} adopts the binarization scheme of Rastegari \etal\cite{rastegariECCV16} as discussed in Sec.~\ref{subsec:training_scheme}.
% A thorough analysis, performed in Sec.~\ref{sec:ablation}, compares different local approximations (binarization methods). 
% In addition to the local and the structural (model design) approximation, the paper studies characteristics of the annotation type of the training data (hand-labeled $\leftrightarrow$ automatic annotations) on the binarization.
The structure of \bDAD{} is given in Fig.~\ref{fig:binarynet}. 

\subsection{Binary Encoder for Feature Extraction}
\label{subsec:BinaryEncoder}
\noindent\textbf{Binary Convolution:}
Without loss of generality, an activation \(A_{l-1} \in \mathbb{R}^{h\times w\times c_i}\) is considered as an input to a convolutional layer \(l \in [1,L]\). In the case of \(l=1\), the activation \(A_1\) is the input image \(I\). Moreover the weights \(W_{l} \in \mathbb{R}^{k\times k\times c_i\times c_o}\) are the trainable parameters of the layer. The sign-function binarizes the real-valued activations \(H_{l-1}\approx \sign(A_{l-1})\). In the inference-stage the weights are considered to \(B\approx \sign(W) \in \{-1,+1\}\). Moreover, the activations are normalized using BatchNormalization \cite{IoffeS15}.
Scale factors \(\alpha\) and \(\beta\), introduced in \cite{rastegariECCV16}, find better estimations for \(W\approx \alpha B\) and \(A\approx \beta H\), see Eq.~\ref{eq:binary_conv}. The first convolutional layer is not binarized due to very few trainable parameters and computations compared to the remaining \bDAD's layers.
\begin{equation} \label{eq:binary_conv}
A_l=\texttt{Conv}(W_S^l,f_S^{l-1})\approx\alpha\beta\texttt{Conv}(B^l,H^{l-1})
\end{equation}

\noindent\textbf{Binary Residual Block:}
The residual block, introduced by He \etal\cite{ResNet_citation}, can be easily binarized learning more complex features as a regular binary convolutional layer, see Sec.~\ref{sec:ablation}. 
In detail, the binary residual block is built of consecutive binary convolutional layers including BatchNormalization and a non-linear activation.
The shortcut connections in binary residual blocks are an adequate technique to overcome the gradient mismatch problem.
Also for the binary version, these blocks combine the information obtained from the previous layer through fusing the identity connections with the output of the current layer. 
%Residual block increase the required on-chip memory and necessary operations due to the extra identity connection.  
%The structure of the binary residual block is given hereinafter. BinaryAct()-BinaryConv()-Relu-BN - … - Add(Bottom,Idetity).

\subsection{Binary Bottleneck with Enlarged Receptive Field}
\label{subsec:BinaryBottleneck}
The bottleneck layers in the segmentation architecture retain the lowest spatial dimensions obtained from the encoder, likewise to an auto-encoder.
Similar to the binary convolutional layer, weights and activations are binarized for the bottleneck, see central building block of Fig.~\ref{fig:binarynet}.

Inspired by DeepLabv3~\cite{10.1007/978-3-030-01234-2_49}, to increase the receptive field of a convolutional layer, dilated convolution introduces zeros to the weights of the respective layer. 
The distance between two neighboring weights is called dilation rate, see also the bottleneck's weights (\textcolor{gray}{gray} parts) of Fig. \ref{fig:binarynet}. 
A typical binary dilated convolution block consists of 1) binarization of the activations, 2) binary convolution, 3) BatchNormalization and 4) non-linear activations such as ReLU. It is important to apply the non-linear activation after the normalization (unit variance and zero mean) to prevent the feature map from losing too much information. Our observation for dilated convolution fits previous investigations for vanilla binary convolution layer \cite{rastegariECCV16,Tang2017HowTT,NIPS2017_6638}.
%Different to transpose convolution\cite{FCN}, which is memory and computationally and memory costly, the dilated (aka. atros) convolution is only computationally expensive during the training stage but more embedded friendly in the deployment. \man{check this statement}
\begin{figure}[ht]
\centering
	\centering
    \subfigure[BPAC based bottleneck.]{\includegraphics[width=0.21\textwidth]{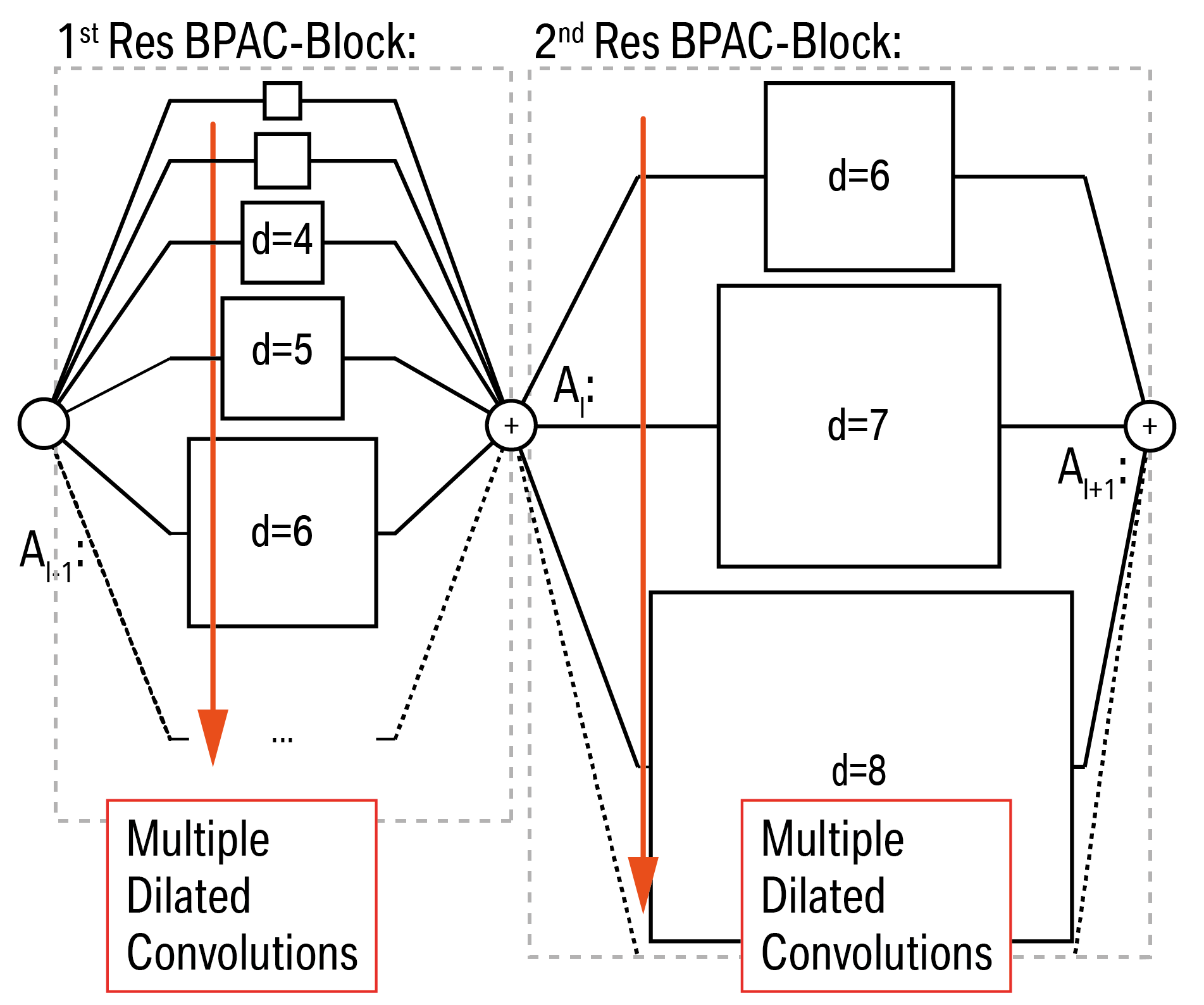}
    \label{fig:bpac_bottle}}
    \subfigure[\BDAD{} bottleneck.]{\includegraphics[width=0.23\textwidth]{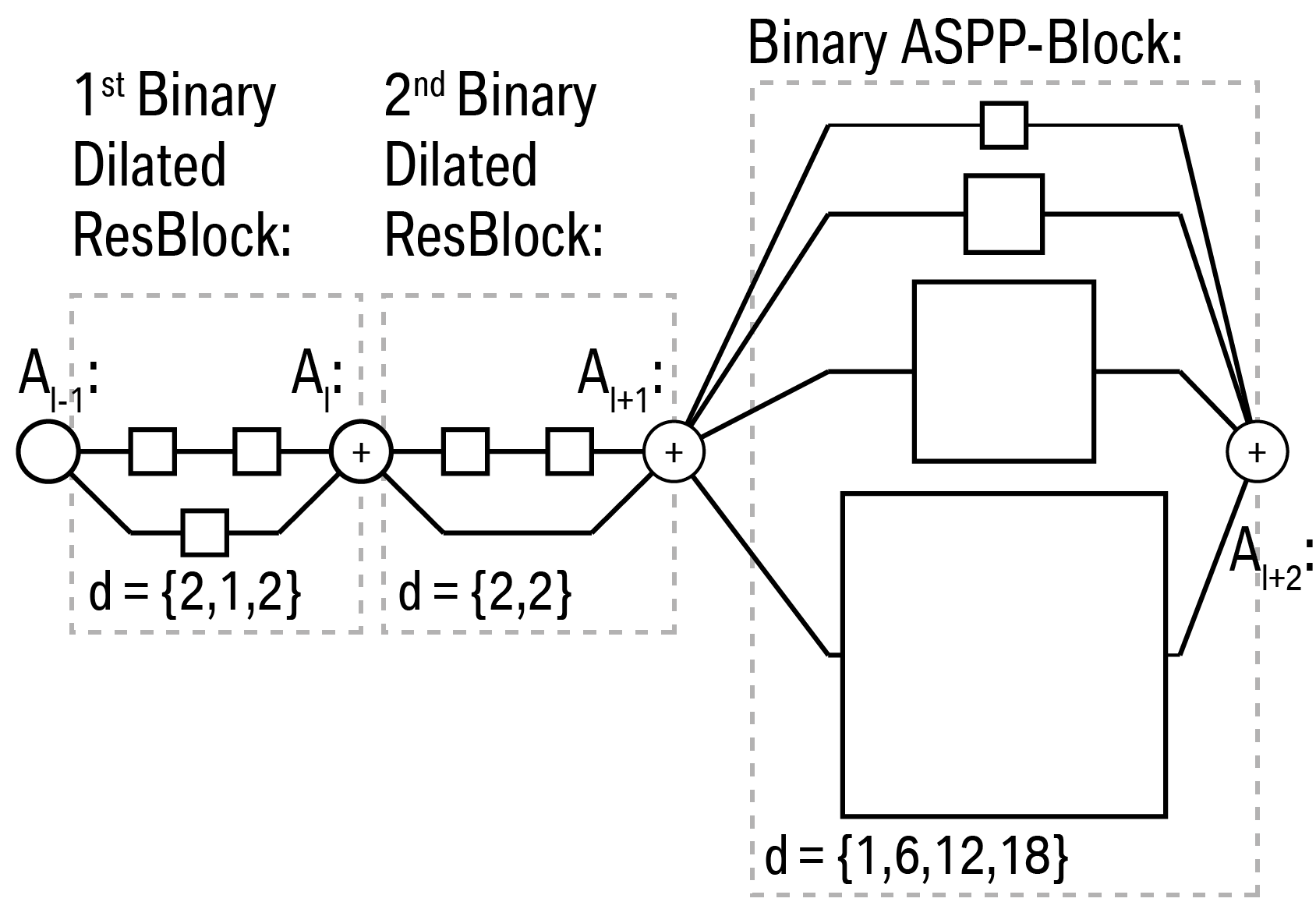}
    \label{fig:bdad_bottle}}
    \caption{Comparison between BPAC~\cite{CVPR19Zhuang3} and BinDAD bottleneck.}
\end{figure}

The central part of the DeepLab~\cite{10.1007/978-3-030-01234-2_49} inspired \bDAD{} is the bottleneck, see Fig.~\ref{fig:bdad_bottle}
In detail, the bottleneck consists of two consecutive binary residual blocks and a binary atrous spatial pyramid pooling (ASPP)-block. The dilation rate $d$=$2$ for the residual blocks and $d=\{1,8,12,18\}$ for the binary ASPP-block is employed. Different to previous residual blocks, the dilated residual blocks do not downsample the feature maps. Thus, the feature resolution of the binary bottleneck is efficiently increased. Upsampling by a factor of 16 instead of 32 is required.

Differently, structured BinaryNet \cite{CVPR19Zhuang3} employs multiple parallel dilated convolution (8), called BPAC-block, in the binary dilated residual blocks. 
Their first residual block has dilation rates $d=\{2,3,4,5,6,7,8,9\}$ and the second one has $d=\{6,7,8,9,10,11,12,13\}$ for GroupNet with 8 bases.
Zhuang \etal\cite{CVPR19Zhuang3} skip the ASPP-block in their design, see also Fig.~\ref{fig:bpac_bottle}.
In case of the DAD task we observe an accuracy degradation with this technique, see Tab.~\ref{tab:ablation_binary_dad_bottleneck}. Moreover, the memory accesses for the BPAC based residual blocks becomes very inefficient.

\subsection{Binary Decoder for Semantic Predictions}
\label{subsec:BinaryDecoder}
Third, to best of our knowledge \bDAD{} is the first work also binarizing the decoder for driveable area detection, see right building block of Fig.~\ref{fig:binarynet}. Employing only binary convolutions enlarges the output of the bottleneck to the size of the original input image $I$ generating pixel-wise predictions for the task of driveable area detection. The binary decoder also consists of bilinear upsampling and a binary score layer.
In detail, after the binary dilated convolution, described in the previous section, linear combination (binary $1\times1$ convolution) of the ASPP feature maps and the encoder skip connection (after the first residual block) is computed. Next, the feature maps are fused in two consecutive binary refinement blocks. The binary refinement blocks consist of $3\times3$ kernels, which is similar to the binary convolutional layer, described above. 
Instead of transpose convolutions, bilinear up-sampling enlarges the feature maps to the size of the input $I$. This is important as the binary transpose convolution would introduce additional operations and would lead to an accuracy degradation, see Sec.~\ref{sec:ablation}. 
% A comparison between the transpose convolution based (FCN) and the dilated convolution based (DeepLab) prediction is carried out, see Tab.~\ref{tab:benchmark_binary_dad}. 

\subsection{Training Scheme of \BDAD}
\label{subsec:training_scheme}
Consider an L-layer BinaryNet (\ie 29 for \bDAD) $f$ which takes $I$ as the input image, trained on semantic labels $Y$, with real-valued latent weights $W\in\mathbb{R}: [-1,1]$  as the trainable parameters. 
$Y$ can refer to expensive handcrafted annotations or to the automatic training data generator (TDG) annotations which are described in the experimental results. The training algorithm is given in Algorithm~\ref{alg:ae_training}.

\begin{algorithm}
\caption{Training an L-layer binary DAD-Net.}
\label{alg:ae_training}
\begin{algorithmic}[1]
\Require a minibatch of images $I$ and lables $Y$, initialized weights $W$ and learning rate $\eta$.
\Statex \emph{Forward propagation:}
\For{$l=2$ to L}
\State Compute $B_l$ and $H_{l-1}$ 
\State $A_l \gets Conv(B_l,H_{l-1})$ \Comment{Eq. \ref{eq:binary_conv}}
\Statex \emph{Optionally:} 
\State $S_l \gets \texttt{MaxPool}(A_l)$
\State $O_l \gets BatchNorm(S_l,\Theta_l)$
\State $A_l \gets ReLu(O_l)$
\EndFor
\State $\Tilde{Y} \gets A_L$
\Statex \emph{Backward propagation}:
\State $gA_L \gets \partial\mathcal{L}/\partial A_L$
\For{$l=L$ to 1}
\State \emph{Optionally:} $g\Theta_l,gO_l \gets BackBN(gA_l,O_l,\Theta_l)$
\State $gW_l, gA_{l-1} \gets BackConv(gS_l,H_{l-1},W_l)$\Comment{Eq.\ref{eq:ste_binary_conv}}
\EndFor

\Statex \emph{Update the trainable parameter}:
\For{$l=1$ to L}
\State $W_l \gets Update(W_l,\eta, gw_l)$\label{step:update}\Comment{Momentum}
\State $\Theta_l \gets Update(\Theta_l,\eta,g\Theta_l)$
\State $\eta \gets \lambda\eta$
\EndFor
\end{algorithmic}
\end{algorithm}
In detail, $B=\sign(W)$ approximates $W$ into the binary domain as $B\in\{-1,+1\}$. 
The gradient of the $\sign$ operation vanishes everywhere and therefore the gradient is estimated in order to update the real-valued weights during the training phase. In the simplest case, the estimated gradient could be obtained by replacing the derivative of $\sign$ with the identity function, see Eq.~\ref{eq:ste_binary_conv}. This is referred as the straight through estimator~\cite{DBLP:journals/corr/BengioLC13}. 
\begin{equation} \label{eq:ste_binary_conv}
g_{w} = g_{W_b} 1_{|w| \leq 1}
\end{equation}

Thus, an efficient set of binary weights $\tilde{B}\in\{-1,+1\}$ is trained by minimizing the expected loss \(\mathcal{L}\) 
according to the prediction $\Tilde{Y}$ and annotations $Y$,
as shown in the Eq.~\ref{eq:train_binary_nets}. $\tilde{B}$ is later used during forward propagation and for the inference on the embedded hardware.
\begin{equation}
\label{eq:train_binary_nets}
\tilde{B} = \underset{W\in[-1, +1]^{L}}{\text{argmin}} \mathcal{L}(\E[(f(I, B)], Y)
\end{equation}

Lines 1-7 show the equivalent forward path of \bDAD, resulting in a semantic prediction $\Tilde{Y}$. The loss is a combination of the pixel-wise cross entropy and a L2 regularization loss. The gradients are computed by minimizing the cost function $\mathcal{L}$ from line 8 to 11 (backpropagation). The STE of Eq.~\ref{eq:ste_binary_conv} is used to estimate the binary weights. 
The gradients $gW$ and $g\Theta$ of the weights $W$ and BatchNorm parameters $\Theta$ are applied by Update(), with Momentum optimizer, see lines 12-15.  
The loss also trains the scaling factors $\alpha$ and $\beta$ associated with weights and the activations~\cite{rastegariECCV16}.

\section{Experiments and Results}
% \nsn{todo for myself: refine this section}
The following section is structured as follows: the datasets, training procedure and performance metrics for the benchmarks of \bDAD{} are introduced in Sec.~\ref{subsec:exp_datasets}; in Sec.~\ref{sec:ablation}, the configuration space of \bDAD{} is explored to get more insight in order to the improve the performance, finally in Sec.~\ref{sec:experiments_benchmakr}, the ~\bDAD{} is compared to SotA semantic segmentation models.

\subsection{Benchmark Datasets and Automatic Annotations}
\label{subsec:exp_datasets}
\noindent\textbf{Cityscapes:}
The CityScapes dataset~\cite{cityscapes_bibtex} consists of 2975 training images, 500 validation images, and 1525 test images including their corresponding ground truth labels. Ground truth labels for the test set are not publicly available. The images of size \(2048\times1024\) show German street scenes along with their pixel-level semantic labels of 30(19) classes. However, for training the raw images are down-sampled to a size of \(1024\times512\). For an elaborate comparison of the ground plane detection, the human labeled ground truth validation data from the road and the parking area class are considered as driveable area. The remaining classes are assigned to the non-driveable area, i.e. pedestrians, sidewalks and cars. 

\noindent\textbf{KITTI Road:}
The KITTI Road dataset~\cite{Geiger2012CVPR} consists of 289 images with manually annotated ground truth labels. This data is split into 259 images for training and 30 images for validation. The raw images are down-sampled using nearest neighbour scaling algorithm from (\(1242\times375\)) to (\(1024\times256\)). The experimental setup for the KITTI Road dataset is similar to the CityScapes dataset. 

\noindent\textbf{Automatic Annotations:}
The training data generator (TDG), proposed by Mayr \etal \cite{Mayr.01.10.201805.10.2018} automatically generates annotations for the task of driveable area segmentation.
% \begin{figure}
%     \centering
%     \subfigure[Manual labels of KITTI.]{\includegraphics[width=4cm]{img/manual_tdg/manual_overlay.png}
%     \label{fig:third_sub}}
%     \subfigure[TDG labels for KITTI.]{\includegraphics[width=4cm]{img/manual_tdg/tdg_overlay.png}
%     \label{fig:third_sub}}
%     \caption{Comparison between Manual and TDG labels for a sample image in KITTI road training set.}
%  \label{fig:tdg_results}
% \end{figure}
% \ccu{todo: do we only use TDG-data, or do we effectively increase the data amount, i.e. complement manual labelling with automatically generated annotations? I adivse for the second and we shall state this explicitly!} 
%\man{point noted. In cases of kitti we do that and now its mentioned in text}
%Although, the data provided by Cityscapes and Kitti Road are impressive and very accurate we would like to investigate how the manual training data influences the performance. 
% The automatic labeling allows to quickly migrate to other datasets that expose technical (resolution of camera, field of view) and environmental (weather, lighting conditions, street types) domain shifts. 
In case of the automatically labeled KITTI data, the data amount is increased compared to manual labeling. 
% However, by using the tdg the training dataset is increased by labeling the images from the odometry benchmark. 
In total, we use automatic annotations generated from 10900 images instead of 259 images as training dataset. 
% This example explains the importance of the TDG data as the training dataset is increased.
This dataset also improves the performance \bDAD{} by providing a good initialization, which is further fine-tuned on the corresponding DAD task.  
% We apply the four step stereo-vision based TDG \cite{Mayr.01.10.201805.10.2018} to label the training data. 
% The system takes image pairs of a stereo camera as input and computes the corresponding disparity map. Then, row-wise histograms of the disparity values are created resulting in the v-disparity map. This mapping allows for an easy determination of the drivable area since it is the major horizontal plane in 3D space. Hence, it can be described by a linear function of disparity $d$ and image row coordinate $v$: $f_{DA}(d,v)$.
% We initialize the linear model based on samples of the bottom area of the image, as the probability of seeing the ground plane there is the highest. Afterwards, we continue to iteratively refine the model in a least-squares fashion by adding samples that satisfy it within a narrow threshold. As soon as the final model is determined, the disparity map can be thresholded in a row-wise manner and every pixel with a disparity that aligns with $f(d,v)$ can be interpreted as being part of the drivable area. -- method
% Finally, the automatic ground plane annotation is denoised and ready to train a BinaryNet.
% As already mentioned, the TDG can be used to automatically label any custom stereo based images as \cite{Mayr.01.10.201805.10.2018} showed that the approach works across datasets and networks. However, in this paper, CityScapes images are used as it allows us to do a proper comparison between manually and automatically labeled data. In the case that we train on TDG labeled data, we only use manually labeled data in the validation phase.

\noindent\textbf{Training Procedure:}
The \bDAD{} is trained using the Momentum optimizer with a base learning rate $\eta=0.01$, the momentum $\gamma=0.9$ and weight decay $\lambda=0.0005$. The learning is dropped by a factor of 0.9 every 8 epochs. For the DAD task, all the models are trained for 240 epochs and the results are reported after retraining the batch statistics.

\noindent\textbf{Performance Metrics:}
The metrics reported in this experiments correspond to mean Intersection-over-Union (IoU), Average Precision (AP), False Positive Rate (FPR) and False Negative Rate (FNR) as used in the KITTI Road challenge \cite{Geiger2012CVPR}.
For applications as autonomous driving, it is crucial that the perception models have real-time capability. 
The modern deep learning inference engines such as NVIDIA-T4 GPU ~\cite{nvidia_t4} and Xilinx FPGAs ~\cite{XilDSP48E1} with DSP48 blocks support SIMD-based bit-wise operations. In particular, a single DSP48 block can perform two 16-bit fixed-point multiplications or 48 XNOR operations at once~\cite{OrthrusPE_DATE}. 
% Thus,24 XNOR operations can be normalized as one fixed point multiplication. 
The normalized compute complexity (NCC), allowing an implementation-wise comparison, is defined as the optimal utilization of MAC and XNOR operations in one compute unit. The DSP48 block serves as a reference implementation to compute NCC for the further experiments.

\subsection{Configuration Space Exploration}
\label{sec:ablation}
The requirement of the following analysis is to determine appropriate modules for \bDAD{} (encoder and decoder), a local binary approximation (XNOR$|$CompactBNN$|$ABC), structural approximation schemes of the bottleneck (BPAC$|$Dilation$|$ASPP) and good initialization scheme(ImageNet$|$Automatic annotations). Analysis is performed on CityScapes dataset.
\noindent\textbf{Encoder/Decoder Selection: }
The Tab.~\ref{tab:ablation_binary_dad_encoder_decoder} compares the mIoU on CityScapes dataset, the number of operations and the memory demand for storing the parameter of different encoder-decoder configurations. A detailed comparison is shown in the supplementary material. The models of Tab.~\ref{tab:ablation_binary_dad_encoder_decoder} are trained in full-precision in order to select the right configuration. Based on a high mIoU, the lowest number of operations and an appropriate memory demand, ResNet18 is chosen as encoder and DeepLabV3 as decoder for the DAD task. Moreover, the standard bottleneck of DeepLabV3 only consists of dilated convolutions and has no transpose convolution in the decoder, enabling an efficient binarization for \bDAD.

\begin{table}[ht]
    \begin{center}
	\caption{Selection of a SotA encoder/decoder configuration.}
	\label{tab:ablation_binary_dad_encoder_decoder}
        \begin{tabular}{llccc}
        \toprule
		\textbf{Encoder} &\textbf{Decoder}& \textbf{mIOU [\%]} & \textbf{GOPs} & \textbf{Mem. [MB]}\\
		\midrule
		VGG16 & FCN8  & 96.75 & 222.1 & 268.5\\
		VGG16 & UNet & 95.92 & 234.0 & 272.2\\
		\textbf{ResNet18} & FCN8 & 97.05 & \textbf{20.93} & 22.60\\
		ResNet18 & UNet & 97.50 & 23.58 & 23.26\\
		\textbf{ResNet18} & \textbf{DeepLabV3} & \textbf{97.54} & 29.77 & \textbf{14.64}\\
        \bottomrule
    	\end{tabular}
        \setlength{\belowcaptionskip}{-12pt}
  \end{center}
  \vspace{-3mm}
\end{table}

\noindent\textbf{Local Binarization Scheme:}
In this section three binarization schemes are analyzed for the DAD task. The results are given in Tab.~\ref{tab:ablation_binary_dad_binarization}. 
By adopting XNOR binarization \cite{rastegariECCV16}, the model is trained with one weight and activation base including scaling factors $\alpha$ and $\beta$. For details see Sec.~\ref{sec:binary_dad}. Differently, CompactBNN \cite{Tang2017HowTT} introduces 3 activation bases and therefore increases the computational complexity of binary convolutional layers. For the DAD task no accuracy improvement is observed. Finally, the binarization scheme of ABC-Net \cite{NIPS2017_6638} introduces multiple activation and weight bases (\ie $3\times3$) to the binary convolution. If only the encoder is binarized with different local approximation schemes, the mIoU remains almost the same (See row. 1, 2, 3). Contrary, if the bottleneck and the decoder are binarized, multiple binarizations result in a degraded driveable area detection. Moreover, NCC and memory demand are increased compared to the XNOR binarization (See row. 4, 5, 6).
\begin{table}[ht]
    \begin{center}
	\caption{Local binary approximation of the encoder and decoder.}
	\label{tab:ablation_binary_dad_binarization}
        \begin{tabular}{l|P{0.25cm}P{0.25cm}c|P{0.7cm}P{0.5cm}P{0.7cm}}
        \toprule
		\multirow{2}{*}{\textbf{Encoder/Decoder}} &\multicolumn{3}{c|}{\textbf{Binarization}} &  \multirow{3}{*}\textbf{mIOU} &{\textbf{NCC}} & \textbf{Mem.}\\
		 & En. & De.& Scheme & [$\%$]&[$\times 10^9$] & [MB]\\
		\midrule
		ResNet18+DeepLab & \cmark& \xmark& \textbf{XNOR} & \textbf{96.93} & 7.30 & 4.83\\
		ResNet18+DeepLab & \cmark& \xmark& Compact & 96.83 & 7.96 & 4.83\\
		ResNet18+DeepLab & \cmark& \xmark& ABC & 96.85 & 9.94 & 5.54\\
		\BDAD &\cmark&\cmark&\textbf{XNOR} &96.23& \textbf{0.73}& \textbf{0.92}\\
		\BDAD &\cmark&\cmark&Compact & 92.40 & 1.96& 0.92\\
		\BDAD &\cmark&\cmark&ABC & 93.26 & 5.65& 2.75\\
        \bottomrule
    	\end{tabular}
        \setlength{\belowcaptionskip}{-12pt}
  \end{center}
  \vspace{-3mm}
\end{table}

\noindent\textbf{Bottleneck Configuration:}
In Tab.~\ref{tab:ablation_binary_dad_bottleneck} different bottleneck configurations are analyzed. 
We replace the bottleneck in BinDAD with the BPAC module proposed in~\cite{CVPR19Zhuang3}. Introducing dilations in the bottleneck improves the mIoU for driveable area detection. Zhuang \etal\cite{CVPR19Zhuang3} argues that BPAC modules capture different object scales making the ASPP obsolete. However, in the case of DAD, a dedicated ASPP block shows better accuracy. Moreover, the implementation of BPAC modules becomes inefficient in HW as the dilation rate increases due to irregular memory access on a general inference processor.

\begin{table}[ht]
    \begin{center}
	\caption{Choosing the bottleneck for \bDAD.}
	\label{tab:ablation_binary_dad_bottleneck}
        \begin{tabular}{lccc}
        \toprule
		\textbf{Bottleneck config.} & \textbf{mIOU} & \textbf{NCC} & \textbf{Mem.}\\
		\midrule
		\BDAD{} w/o Dilations + ASPP &94.80&0.65&  0.69\\
        \BDAD{} w \textbf{Dilations} w/o ASPP &95.18&\textbf{0.65}& \textbf{0.69}\\
        \BDAD{} w BPAC~\cite{CVPR19Zhuang3}&96.02& 1.36 & 0.69\\
        \BDAD{} w \textbf{Dilations w ASPP}&\textbf{96.23}&0.73 & 0.92\\
        \bottomrule
    	\end{tabular}
        \setlength{\belowcaptionskip}{-12pt}
  \end{center}
  \vspace{-3mm}
\end{table}

\textbf{Automatic annotations } 
In the field of AD, automatic annotations are a low priced alternative to costly hand labeled data enabling a competitive deployment. On the one hand, automatic annotations are noisy and have much higher variance than manually labeled finite datasets, e.g. CityScapes. On the other hand, BinaryNets are prone to a degradation in accuracy because of their limited learning capabilities. When \bDAD{} is trained on automatic annotations, it achieves on-par accuracy compared to SotA semantic segmentation model, \ie DeepLab. Referred to Tab.\ref{tab:binDAD_tdg_kitti}, TDG data have an mIoU of $69.7\%$ with respect to the manually labeled validation dataset of KITTI (See row. 1). The training of DeepLabV3 on only automatic annotations (TDG) results in an mIoU of (86.1\%). The $38\times$ more efficient \bDAD{} achieves on-par accuracy (See row. 2, 3).

\begin{table}[ht]
    \begin{center}
	\caption{\BDAD 's performance on automatic annotations.}
    \label{tab:binDAD_tdg_kitti}
        \begin{tabular}{llp{0.6cm}p{0.6cm}p{0.6cm}p{0.6cm}}
        \toprule
		\textbf{Model} & \textbf{Taining Data} &\textbf{mIOU} & \textbf{Acc.} \\
		\midrule
		TDG \cite{Mayr.01.10.201805.10.2018} & TDG &69.70 &87.73 \\
		DeepLabV3&TDG &86.11&90.49\\
        \BDAD &TDG &85.33&90.49 \\
        \BDAD &CityScapes & 96.23 &  96.13  \\
        \BDAD &\textbf{TDG + CityScapes} &\textbf{96.60}&\textbf{96.68} \\
        \bottomrule
    	\end{tabular}
        \setlength{\belowcaptionskip}{-12pt}
  \end{center}
  \vspace{-3mm}
\end{table}
\begin{table*}[ht]
    \begin{center}
	\caption{Performance comparison of \bDAD to baseline models. Computations refers to MAC and XNOR operations. The NCC metric (column 5) considers an efficient HW implementation,e.g. with SIMD-based acceleration of XNOR ops.}
	\label{tab:benchmark_binary_dad}
        \begin{tabular}{llcccccccc}
        \toprule
		\multirow{2}{*}{\textbf{Approach}} & \multirow{2}{*}{\textbf{Datasets}} & \textbf{Parameters} & \textbf{Computations} &\textbf{NCC}& \textbf{mIOU} &\textbf{Accuracy}& \textbf{FPR}& \textbf{FNR}\\
		 &  & [MB] & [GOP] & $\times 10^9$ & [\%] &[\%] & [\%] & [\%] \\
		\midrule
        FCN8s  &  CityScapes &   22.60 & 20.93 & 10.47& 96.94& 97.01 &1.62& 5.71\\
        DeepLabv3 &  CityScapes & 14.63 & 26.54 & 14.89 &97.30&97.29&\textbf{1.32}& 5.41\\
        UNet  &  CityScapes & 23.26 & 23.58 &11.79&\textbf{97.50}&\textbf{97.55}&1.59&\textbf{5.33}\\
        FCN8s-XNOR  & CityScapes & 1.41  & 20.93 & \textbf{0.66} & 95.19 & 95.35 & 2.72 & 8.54\\
        % FCN8s-ABCNet  &  CityScapes & & &-&-\\
        \textbf{\BDAD{} (Ours)}  &  \textbf{CityScapes+TDG} &\textbf{0.92}& 29.77 &0.73 & 96.60 &  96.68 &1.96 &  6.16 \\
		\midrule
		FCN8s  &  KITTI Road & 22.60  &10.50 &5.250 & \textbf{95.43}& \textbf{97.71}&3.92&1.91 \\
        DeepLabv3 &  KITTI Road & 14.63  & 20.93 & 10.46 & 94.45& 96.34&\textbf{1.02}&2.12\\
        UNet  &  KITTI Road &23.26 &10.71&5.36&93.26&95.50&4.81&\textbf{1.51}\\
        FCN8s-XNOR  &  KITTI Road &  1.41 & 10.50& 0.33 & 92.10& 96.34&4.92&3.44 \\
        % FCN8s-ABCNet  &  Kitti Road & & &-&-\\
        \textbf{\BDAD{} (Ours)}  &  KITTI Road&\textbf{0.92} & 13.26 &\textbf{0.27} & 95.25 &97.05& 7.83 & 1.82 \\
		%BinDAD-ABC  &  Kitti Road & & &-&-\\
		\bottomrule
    	\end{tabular}
        \setlength{\belowcaptionskip}{-12pt}
  \end{center}
\end{table*}
Second, the training of BinaryNets is highly sensitive to initialization. In~\cite{deeplabv3plus2018}, the encoder is pre-trained on ImageNet and later fine-tuned on segmentation task. With the same initialization strategy, \bDAD{} achieves an mIOU of (96.23\%). However, when \bDAD{} is fine-tuned on the pre-trained TDG data, there is an improvement of 0.37\% in mIOU. (See, row 4, 5) in Tab.\ref{tab:binDAD_tdg_kitti}. 

\subsection{Comparison with the State-of-the-Art}
\label{sec:experiments_benchmakr}

In this section, the structure of the proposed \bDAD{} is analyzed and evaluated against SotA models for semantic segmentation on both public datasets. Also a variety of encoder models with different binarization methodologies proposed in~\cite{rastegariECCV16}, \cite{Tang2017HowTT} and~\cite{NIPS2017_6638} are employed. 
\begin{table}[ht]

    \begin{center}
	\caption{A dedicated two class DAD model (right) is compared to a multi-class model (left) justifying \bDAD.
% 	Alongside with a good mIoU, semantic segmentation models for the task of DAD require a high recall and a high precision. A multi-class model (left) is not able to perform equally good over all three metrics. Moreover, two class binary DAD performs on par results compared to its full precison counterpart (right). -- Included in the text (Baseline comparison)
	}
	\label{tab:19_2_class_comparision}
        \begin{tabular}{p{1.8cm}|p{1cm}p{1.2cm}|p{1cm}p{1.2cm}}
        \toprule
         & \multicolumn{2}{c}{\textbf{All 19 classes}} & \multicolumn{2}{|c}{\textbf{Only DAD (2 class)}}\\
         \midrule
		\multirow{2}{*}{\textbf{Metric\textbackslash Model}} & \textbf{FP} & \textbf{Binary} & \textbf{FP}& \textbf{Binary}\\
		 & \textbf{DeepLab} & \textbf{DAD-Net} & \textbf{DeepLab}& \textbf{DAD-Net}\\
		\midrule
		\textbf{Road IoU [\%]}  & \multicolumn{1}{c}{97.23} & \multicolumn{1}{c|}{97.22}& \multicolumn{1}{c}{\textbf{97.55}}& \multicolumn{1}{c}{96.79}\\
		\textbf{meanIoU [\%]} & \multicolumn{1}{c}{63.43} & \multicolumn{1}{c|}{58.12} & \multicolumn{1}{c}{\textbf{97.30}} & \multicolumn{1}{c}{96.60}\\
		\textbf{Precision [\%]} & \multicolumn{1}{c}{79.23} & \multicolumn{1}{c|}{80.00} & \multicolumn{1}{c}{\textbf{97.36}} & \multicolumn{1}{c}{94.30}\\
		\textbf{Recall [\%]} & \multicolumn{1}{c}{71.12} & \multicolumn{1}{c|}{71.21} & \multicolumn{1}{c}{\textbf{95.13}} & \multicolumn{1}{c}{93.53}\\
		\bottomrule
    	\end{tabular}
        \setlength{\belowcaptionskip}{-12pt}
  \end{center}
  \vspace{-3mm}
\end{table}

% \begin{table*}[ht]
% \label{tab:19_2_class_comparisopn}
%     \begin{center}
% 	\caption{Motivation for a dedicated drivable area segmentation. Alongside with a good mIoU, semantic segmentation models for the task of DAD require a high recall and a high precision. A multi-class model (left) is not able to perform equally good over all three metrics. Moreover, two class binary DAD performs on par results compared to its full precision counterpart (right).}
%         \begin{tabular}{l|cc|cc}
%         \toprule
%          & \multicolumn{2}{c}{\textbf{All 19 classes}} & \multicolumn{2}{|c}{\textbf{Only Drivable Area Segmentation}}\\
%          \midrule
% 		\multirow{2}{*}{\textbf{Class\textbackslash Model}} & \textbf{Full Precision} & \textbf{Binary} & \textbf{Full Precision}& \textbf{Binary}\\
% 		 & \textbf{ResNet + FCN} & \textbf{XNORNet + BinaryFCN} & \textbf{ResNet + FCN}& \textbf{BinaryDAD-Net}\\
% 		\midrule
%         road  & 98.175 & 97.668& 96.953& 96.080\\
%         sidewalk & 84.254 & 82.9855&-&-\\
%         traffic light  & 43.182 & 20.993&-&-\\
%         terrain  & 61.618 & 65.141&-&-\\
%         ...&...&...&...&...\\
%         bicycle  &70.823 & 62.248&-&-\\
%         others &-&-&97.378&97.448\\
% 		\midrule
% 		\textbf{meanIU} & 63.424 & 58.335&97.167&\textbf{96.764}\\
% 		\midrule
% 		\textbf{Precision} & 0.80& 0.80&0.95&\textbf{0.95}\\
% 		\textbf{Recall} &0.71 &0.70&0.97&\textbf{0.96}\\
% 		\bottomrule
%     	\end{tabular}
%         \setlength{\belowcaptionskip}{-12pt}
%   \end{center}
% \end{table*}
Firstly, a justification of a dedicated two class driveable area detection is given in Tab.~\ref{tab:19_2_class_comparision}. The IoU of the individual road class of the full-precision segmentation network is 97.23\%. However, training a model with many classes, i.e. 19, causes the precision and recall for the task of driveable area segmentation to degrade compared to the dedicated two class DAD task.
This also holds for the binary implementation, see column 2 and 4.
Moreover, \bDAD{} achieves comparable results (only -0.70\% mIoU) compared to the bulky full-precision DeepLabV3 on the DAD task.
\begin{figure}
    \begin{center}
    % {\includegraphics[width=0.18\textwidth]{img/fig3/label/overlay_label_iter_0.png}
    % \label{fig:rawimage}}
    {\includegraphics[width=0.15\textwidth]{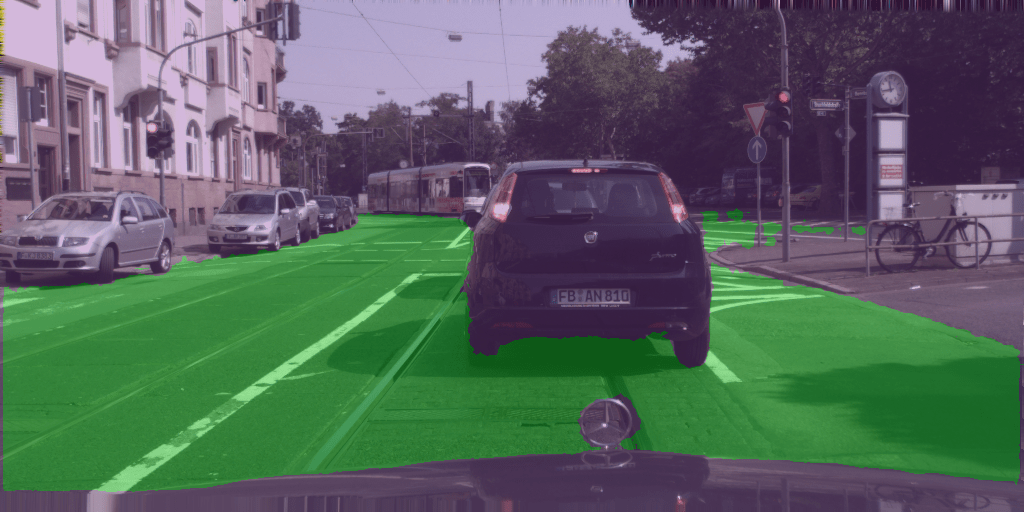}
    \label{fig:rawimage}}
    {\includegraphics[width=0.15\textwidth]{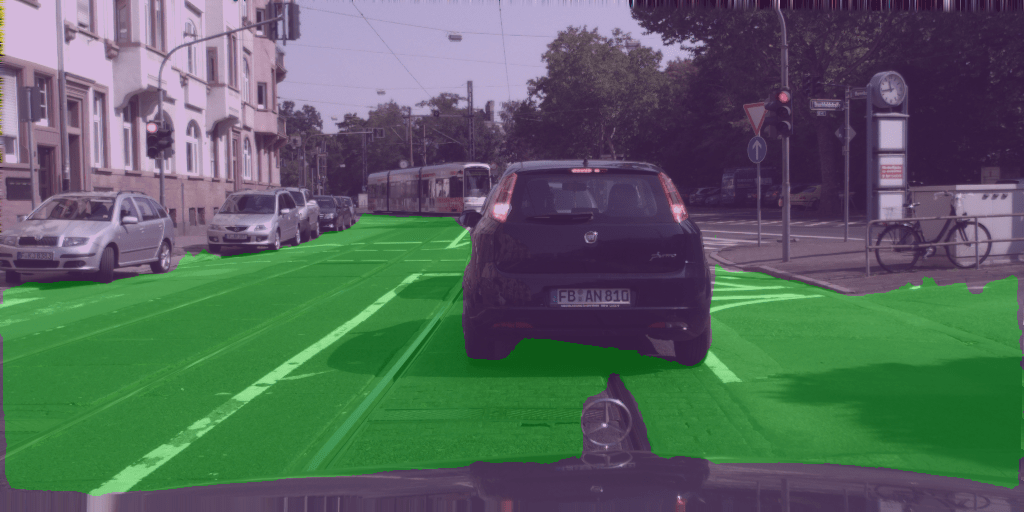}
    \label{fig:rawimage}}
    % {\includegraphics[width=0.18\textwidth]{img/fig3/benchmark3/overlay_iter_0.png}
    % \label{fig:rawimage}}
    {\includegraphics[width=0.15\textwidth]{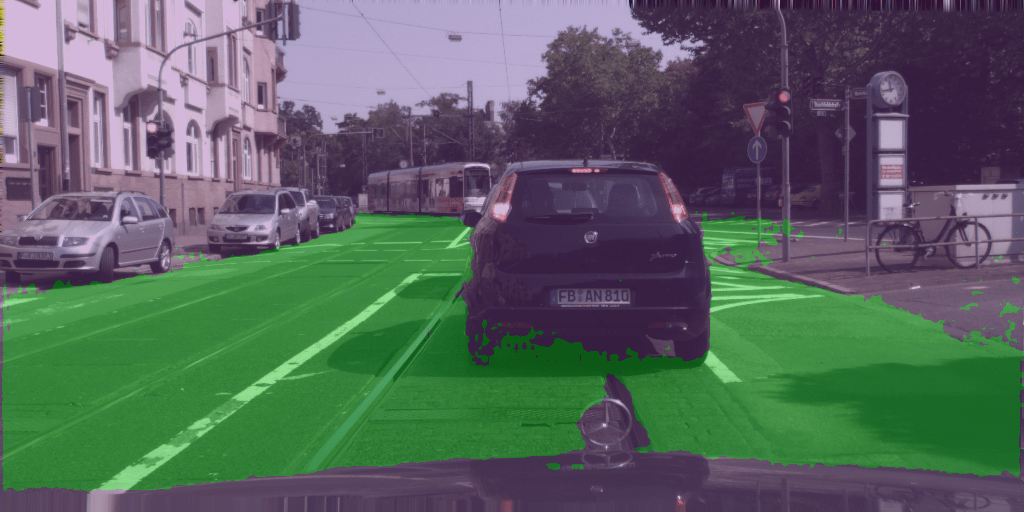}
    \label{fig:rawimage}}
    \end{center}
    \begin{center}
    % {\includegraphics[width=0.18\textwidth]{img/fig3/label/overlay_label_iter_1.png}
    % \label{fig:rawimage}}
    {\includegraphics[width=0.15\textwidth]{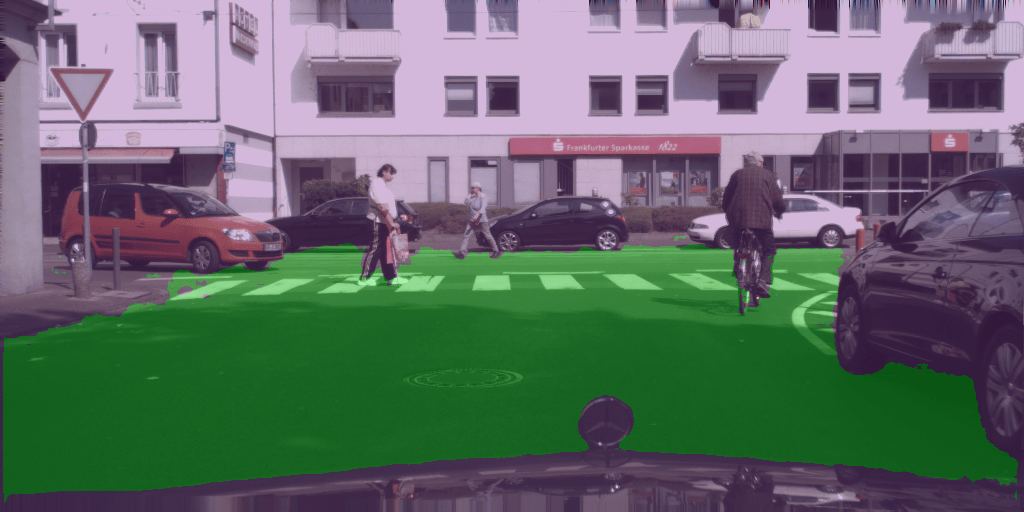}
    \label{fig:rawimage}}
    {\includegraphics[width=0.15\textwidth]{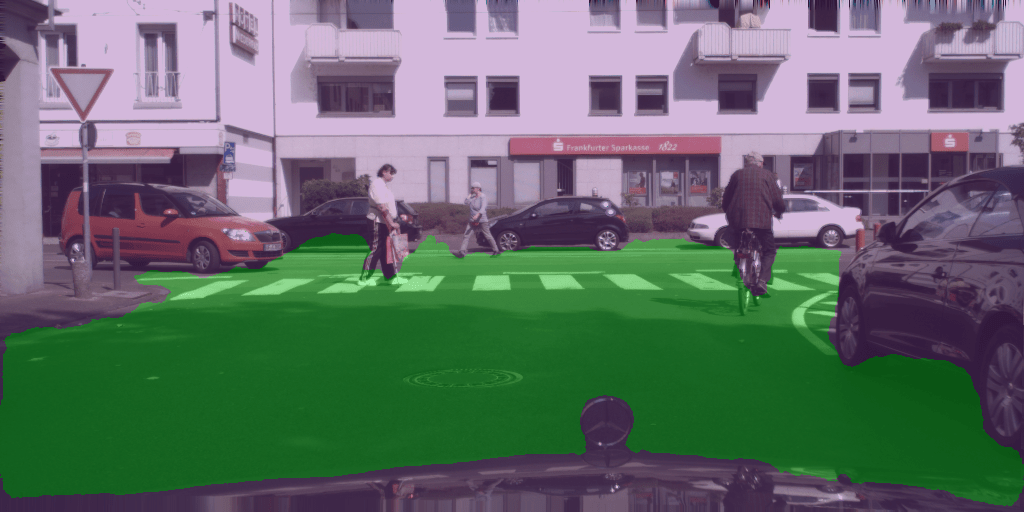}
    \label{fig:rawimage}}
    % {\includegraphics[width=0.18\textwidth]{img/fig3/benchmark3/overlay_iter_1.png}
    % \label{fig:rawimage}}
    {\includegraphics[width=0.15\textwidth]{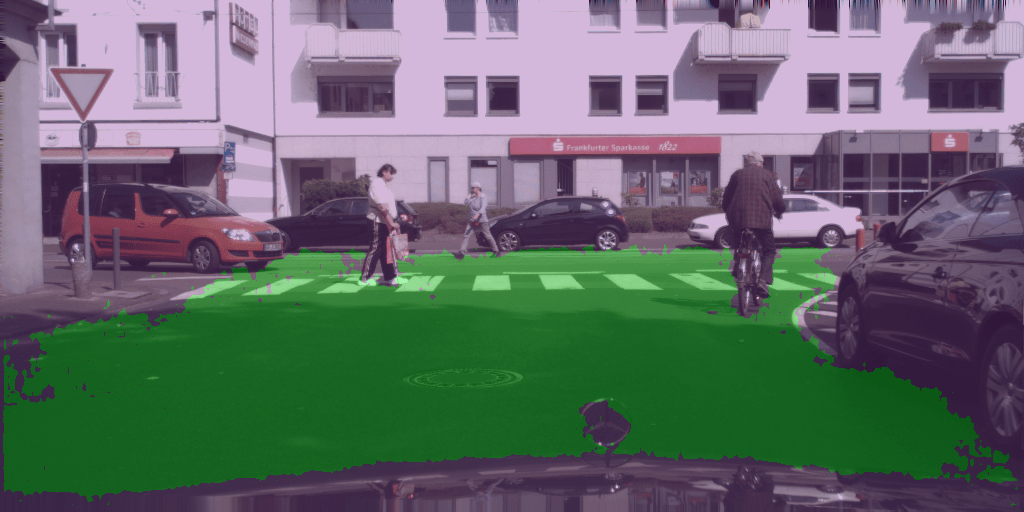}
    \label{fig:rawimage}}
    \end{center}
    \begin{center}
    % \subfigure[Manual Labels~\cite{cityscapes_bibtex}]{\includegraphics[width=0.18\textwidth]{img/fig3/label/overlay_label_iter_3.png}
    % \label{fig:rawimage}}
    \subfigure[DeepLabv3~\cite{10.1007/978-3-030-01234-2_49}.]{\includegraphics[width=0.15\textwidth]{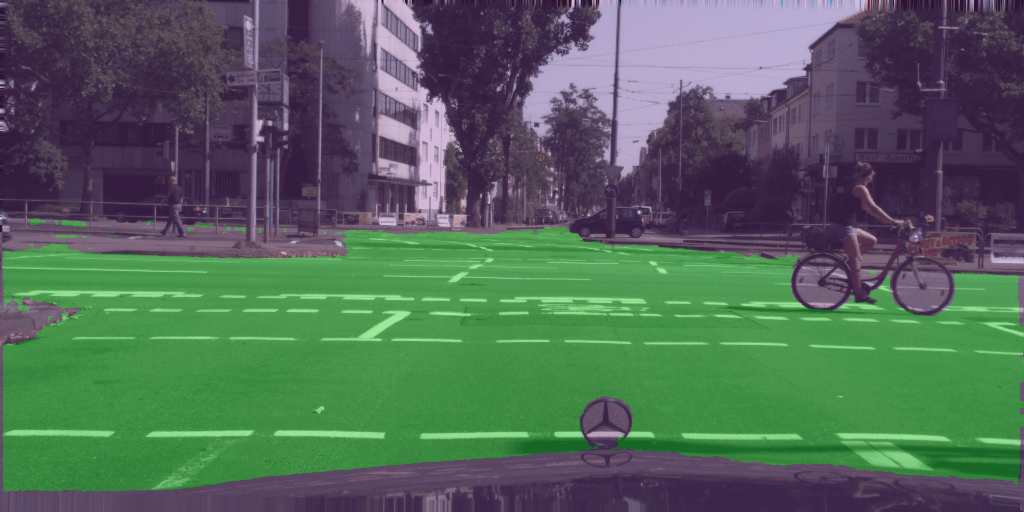}
    \label{fig:rawimage}}
    \subfigure[FCN8s~\cite{Shelhamer:2017:FCN:3069214.3069246}.]{\includegraphics[width=0.15\textwidth]{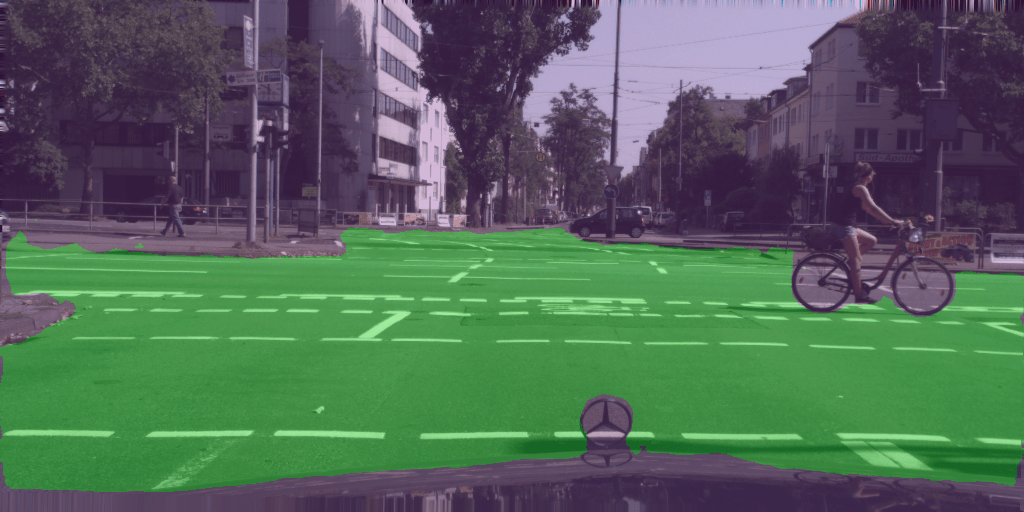}
    \label{fig:rawimage}}
    % \subfigure[FCN8s-XNOR~\cite{rastegariECCV16}]{\includegraphics[width=0.18\textwidth]{img/fig3/benchmark3/overlay_iter_3.png}
    % \label{fig:rawimage}}
    \subfigure[\BDAD.]{\includegraphics[width=0.15\textwidth]{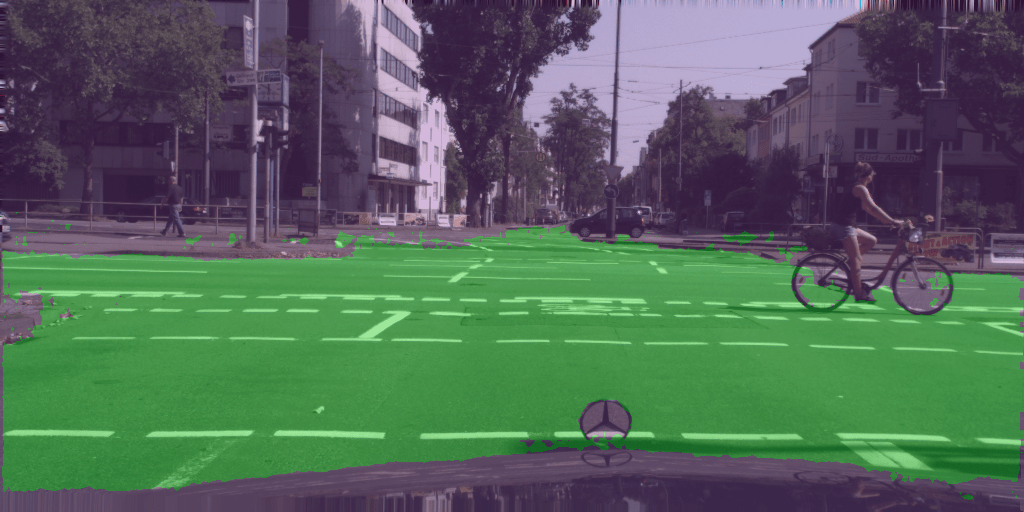}
    \label{fig:rawimage}}
    \end{center}
    
    \caption{Quantitative results on different scenarios in CityScapes dataset. The last column shows the semantic predictions of \bDAD.}
    \label{fig:sample_subfigures}
\end{figure}

Full-precision models have been proven to be dispensable for most prediction tasks, rendering 16-bit fixed-point representations an adequate alternative \cite{mixed_precision_train}. With respect to the memory requirements and the compute complexity, \bDAD{} is compared to a 16-bit implementation rather than a 32-bit one. Tab.~\ref{tab:benchmark_binary_dad} shows the performance of different CNNs for driveable area detection, including \bDAD{}. The models are trained on the CityScapes and the KITTI Road datasets separately. \BDAD{} achieves a mIoU of $96.60\%$ on the CityScapes and $95.25\%$ on the KITTI Road dataset which constitutes to an improvement of $+1.05\%$ ($3.15\%$) compared to the previous best BinaryNet.
The improvement over FCN8s-XNOR is due to the highly representational bottleneck block discussed in Sec.~\ref{subsec:BinaryBottleneck}. Apart from the comparison with BinaryNets, \bDAD{} observes a slight accuracy degradation of -$0.9$ (-$0.7$) compared to the state-of-the-art full-precision segmentation networks, i.e. UNet (DeepLabV3). Fig. \ref{fig:sample_subfigures} displays some quantitative results on different scenarios in the CityScapes dataset. The misclassified pixels of \bDAD{}, compared to the full-precision counterpart, can be easily regularized by a ground plane detection algorithm.
Moreover, the predictions show that a minor accuracy degradation is negligible taking the performance advantages into consideration.

\section*{Conclusion}
This paper introduces a novel binary driveable area detector (binary DAD-Net) required in the field of autonomous driving. \BDAD{} is fully binarized, including the encoder, the bottleneck and the decoder. 
An elaborate study is performed to explore various components of \BDAD, namely the model structure, the binarization scheme and the ground-truth annotations for training.
Along with automatically generated training data, binary DAD-Net achieves state-of-the-art semantic segmentation results $96.60\%$(-$0.7\%$) on the CityScapes dataset.
The proposed driveable area detector is very memory efficient, with only 0.9MB parameters (-$15.9\times$). Moreover, \BDAD{} shows its superior performance w.r.t. an embedded implementation, by drastically reducing the computational complexity ($14.3\times$) compared to previous work.

\bibliography{IEEEabrv,IEEEexample}
%\appendix
%\newpage
%\input{supplementary/supplementary.tex}
\end{document}